\documentclass[journal]{IEEEtai}

\usepackage[colorlinks,urlcolor=blue,linkcolor=blue,citecolor=blue]{hyperref}

\usepackage{color,array}

\usepackage{graphicx}

\usepackage[table]{xcolor}

\usepackage{cite}

\usepackage{amsmath}
\usepackage{booktabs, tabularx}
\usepackage{multirow}
\usepackage{multicol}
\usepackage{threeparttable}
\usepackage{longtable}
\usepackage{threeparttablex}
\usepackage{algorithm}
\usepackage{algorithmic}
\usepackage{mathtools}
\usepackage{amssymb}

    \newcommand{\midsepremove}{\aboverulesep = 0mm \belowrulesep = 0mm}
    \midsepremove
    \newcommand{\midsepdefault}{\aboverulesep = 0.1mm \belowrulesep = 0.1mm}
    \midsepdefault
    
 \newcommand{\notesize}{\fontsize{6.8}{12}\selectfont}

\begin{document}

\title{Comprehensive Attribute Encoding and Dynamic LSTM HyperModels for Outcome Oriented Predictive Business Process Monitoring}

\author{Fang Wang, Paolo Ceravolo, \IEEEmembership{Member, IEEE} and Ernesto Damiani, \IEEEmembership{Senior Member, IEEE},
\thanks{Fang Wang (Florence Wong) is with College of Computing and Mathematical Sciences, Khalifa University, Abu Dhabi, UAE (e-mail: florence.wong@ku.ac.ae).}\thanks{Paolo Ceravolo is with Computer Science Department, University of Milan, Milan, Italy
 (e-mail: paolo.ceravolo@unimi.it)}\thanks{Ernesto Damiani is with Center for Cyber-Physical Systems  Khalifa University, Abu Dhabi, UAE, and with College of Computing and Mathematical Sciences Khalifa University, Abu Dhabi, UAE, (e-mail: ernesto.damiani@ku.ac.ae).}\thanks{\textbf{Disclaimer}: This manuscript is a preprint currently under review at IEEE Transactions on Artificial Intelligence (IEEE TAI). It has not yet undergone peer review or been accepted for publication. Please do not use this version to assess the final scientific validity of the work.}
 \thanks{Code repository: https://github.com/skyocean/HyperLSTM-PBPM}
}

\maketitle

\begin{abstract} Predictive Business Process Monitoring (PBPM) aims to forecast future outcomes of ongoing business processes. However, existing methods often lack flexibility to handle real-world challenges such as simultaneous events, class imbalance, and multi-level attributes. While prior work has explored static encoding schemes and fixed LSTM architectures, they struggle to support adaptive representations and generalize across heterogeneous datasets. To address these limitations, we propose a suite of dynamic LSTM HyperModels that integrate two-level hierarchical encoding for event and sequence attributes, character-based decomposition of event labels, and novel pseudo-embedding techniques for durations and attribute correlations. We further introduce specialized LSTM variants for simultaneous event modeling, leveraging multidimensional embeddings and time-difference flag augmentation. Experimental validation on four public and real-world datasets demonstrates up to 100\% accuracy on balanced datasets and F1 scores exceeding 86\% on imbalanced ones. Our approach advances PBPM by offering modular and interpretable models better suited for deployment in complex settings. Beyond PBPM, it contributes to the broader AI community by improving temporal outcome prediction, supporting data heterogeneity, and promoting explainable process intelligence frameworks.

\end{abstract}

\begin{IEEEImpStatement}
Business processes underpin daily operations across healthcare, finance, public services, and logistics. Predicting the outcome of ongoing processes—such as whether a loan will be approved or a shipment delayed—can save time, reduce costs, and improve service. However, current AI tools often struggle with real-world complexity, like overlapping events or imbalanced data. Our work introduces adaptive, interpretable models that overcome these hurdles, making accurate predictions in more realistic settings. This has the potential to enhance transparency and decision-making in mission-critical systems, reduce process inefficiencies, and support timely interventions. The modular design also facilitates integration into existing systems, promoting technological uptake across sectors. By aligning AI capabilities with real-world business demands, this research helps bridge the gap between academic innovation and practical impact—advancing both the science and application of trustworthy, human-centric AI.
\end{IEEEImpStatement}


\section{Introduction}
Predictive Business Process Monitoring (PBPM) has emerged as a critical application of Artificial Intelligence(AI), leveraging machine learning to forecast process outcomes based on event log data \cite{10418930}. However, while deep learning models have shown promising results in event-level predictions, sequence-level outcome prediction remains fundamentally limited by three major AI challenges: 1) capturing complex interdependencies between event attributes, sequence-level characteristics, and temporal dynamics; 2) lack of adaptive learning mechanisms, limiting model generalization across diverse datasets; 3) encoding and representation bottlenecks, leading to information loss in heterogeneous event logs.

Most existing outcome-oriented PBPM models rely on traditional machine learning techniques, such as decision trees, clustering, and ensemble methods \cite{de2016general,di2016clustering,leontjeva2015complex,lakshmanan2011predictive}. However, these methods fail to capture the sequential dependencies inherent in event logs, making them unsuitable for tasks requiring deeper temporal understanding. An additional challenge in PBPM is the heterogeneous and dynamic nature of event and sequence attributes, particularly temporal features. Overlapping events with varying completion times further complicate the modeling process \cite{10.1007/978-3-031-61057-8_5}, necessitating more robust encoding and embedding techniques to effectively extract meaningful patterns ~\cite{tavares2023trace}. This phenomenon of simultaneous or temporally-aligned events is frequently observed in real-world domains such as healthcare, logistics, and service management, where parallel activities occur by design. Failing to model this properly can degrade performance and reduce interpretability in predictive tasks. While Long Short-Term Memory (LSTM) networks have shown promise for capturing long-range dependencies and temporal structures—particularly for tasks like next-event prediction and remaining-time estimation \cite{evermann2017predicting,tax2017predictive}—their effectiveness in outcome prediction remains limited. This is primarily due to their dependence on encoding strategies, which, if poorly designed, can lead to information loss or suboptimal feature representation. Unlike natural language processing, where standardized architectures can generalize across datasets, PBPM datasets exhibit significant structural variability, requiring adaptive modeling approaches.

To address these challenges, we propose a comprehensive framework that integrates novel encoding and embedding strategies and self-tuning LSTM hypermodels for outcome prediction. This work makes three key contributions to PBPM: (1) We propose novel encoding and embedding strategies tailored for event logs, including linguistic decomposition of event labels and pseudo-embedding techniques that capture attribute correlations and dynamically bin event durations. (2) We tackle the challenge of simultaneous events through multidimensional embeddings and time-difference label augmentation, ensuring robust representation of temporal relationships. (3) We design a suite of dynamic LSTM hypermodels—B-LSTM, D-LSTM, DC-LSTM, and T-LSTM—each incorporating self-tuning hyperparameters to adapt to diverse datasets. These contributions represent a significant advancement in PBPM, offering a novel and adaptive framework that addresses key limitations of existing approaches. Experimental results demonstrate the effectiveness of our framework in improving outcome prediction accuracy.

The remainder of this paper is organized as follows. Section \ref{sec:RW} reviews related work in PBPM, highlighting gaps in existing approaches. Section \ref{sec:AE} introduces our attribute encoding and embedding strategies, while Section \ref{sec:HM} and  \ref{sec:EP} details the architectures of the proposed LSTM hypermodels and experiments, and Section \ref{sec:RE} and \ref{sec:CC} concludes with a summary of findings and future research directions.

\section{Related Work}
\label{sec:RW}
Accurately predicting the final outcome of an ongoing business process instance—such as loan approvals or process deviations—is a critical yet underexplored task in Predictive Business Process Monitoring (PBPM) \cite{maggi2014predictive,pika2016evaluating, di2016clustering,teinemaa2019outcome,genga2019predicting,di2022predictive}. While PBPM has traditionally focused on next activity prediction \cite{evermann2017predicting} and remaining time estimation \cite{tax2017predictive}, sequence-level outcome prediction remains underexplored and highly challenging. Early PBPM research relied on symbolic sequence classification, where classifiers were trained on manually engineered features extracted from event logs \cite{leontjeva2015complex,pasquadibisceglie2020orange,santoso2018specification}. Popular models include Decision Trees (DT) \cite{grigori2001improving,grigori2004business,castellanos2005predictive}, Random Forest (RF) \cite{leontjeva2015complex}, XGBoost \cite{senderovich2017intra}, and Support Vector Machines (SVMs) \cite{kang2012periodic}. While RF and boosting-based models have shown robust performance in structured datasets, they fail to capture temporal dependencies, event correlations, and sequence-level interactions, leading to suboptimal generalization \cite{teinemaa2019outcome}. For example, SVM approach in \cite{kang2012periodic} demonstrated 82\% accuracy in lab conditions but dropped to 63\% when applied to actual hospital workflows \cite{rama2021deep}.

To address the limitations of manual feature engineering, Recurrent Neural Networks (RNNs) and, more specifically, Long Short-Term Memory (LSTM) networks have been applied to PBPM tasks \cite{rama2021deep}. Initial studies focused on next-event prediction \cite{evermann2017predicting,schonig2018deep} and remaining time estimation \cite{navarin2017lstm,tax2017predictive,camargo2019learning}. Subsequent improvements introduced hierarchical attention mechanisms \cite{jalayer2022ham}, time-aware modeling \cite{nguyen2021time}, and multi-attribute event representations \cite{lin2019mm}. Despite these advances, LSTMs still face the following fundamental challenges in accurately predicting sequence-level outcomes, particularly in capturing complex dependencies and ensuring model adaptability:

\begin{itemize}
\item \textbf{Encoding bottlenecks}: The effectiveness of LSTM models in PBPM depends heavily on how events, sequences, and attributes are encoded and integrated into the model architecture \cite{rama2021deep,Harane2020,verenich2019survey}. Traditional one-hot and frequency-based encoding strategies effectively capture attribute independence but often overlook latent relationships between attributes \cite{tavares2023trace}.  While some studies have explored word embeddings inspired by natural language processing \cite{pegoraro2021text}, these methods fail to adequately represent the heterogeneous and hierarchical structure of business process event logs. 

\item \textbf{Limited handling of overlapping events}: Many business processes include simultaneous events, which are not effectively represented in standard sequence models \cite{neu2022systematic}.

\item \textbf{Lack of adaptive learning}: Unlike NLP models that generalize across datasets, PBPM datasets exhibit significant variability in event structures and attribute distributions \cite{teinemaa2016predictive, metzger2014comparing}. Most existing LSTM models rely on manually tuned hyperparameters, limiting their generalization capabilities \cite{lin2019mm}. While some studies explored auto-tuning \cite{di2018genetic}, none have tackled outcome prediction with self-tuning LSTMs.

\item \textbf{Performance instability across datasets}: Prior studies on sequence classification \cite{wang2019outcome, folino2019learning, hinkka2019classifying, pegoraro2021text} have struggled with low accuracy and fail to account for dynamic interactions between event and sequence level attributes.
\end{itemize}

Given these challenges, there is a clear need for an LSTM-based framework that not only improves event representation but also enhances model adaptability and robustness across diverse PBPM datasets. To address this, we propose a series of LSTM hypermodels (B-LSTM, D-LSTM, DC-LSTM, and T-LSTM) specifically designed for outcome prediction in PBPM. Our framework extends existing encoding and embedding techniques by introducing attribute correlation pseudo-embeddings and time-difference label augmentation, improving the expressiveness of input representations while effectively handling overlapping events in trace logs. Additionally, our models incorporate self-tuning hyperparameters, ensuring adaptability across diverse PBPM datasets. By tackling these challenges, our framework significantly advances PBPM by improving event representation, enhancing model adaptability, and ensuring robustness across diverse datasets—ultimately enabling more reliable and accurate outcome predictions in real-world applications.

\section{Attribute Encoding and Embedding}
\label{sec:AE}
\subsection{Attribute Notation and Timestep Definition}
In process based sequence data, attributes are analysed across two hierarchical levels: the event level and the sequence level. Let $X_i$ denote a single event and $S_j$ the entire sequence, where $X_i \in S_j$ indicates that event $X_i$ is part of sequence $S_j$. At the event level, attributes ($F_i$) include universal attributes ($U_i \subset F_i$), such as event type and sub-status—denoted as $U^a_i$, $U^b_i$, $U^c_i$, etc.—as well as time-related features: start time ($T^s_i$), end time ($T^c_i$), and duration ($T^d_i = T^c_i - T^s_i$). These temporal features enable the identification of sequences containing simultaneous or overlapping events. Specific attributes ($B_i \subset F_i$)—denoted as $B^a_i$, $B^b_i$, etc.—are applicable only to certain event types.

At the sequence level, attributes (e.g., $H^a_j$, $H^b_j$) describe holistic properties of the sequence $S_j$, such as case category or total trace duration.

Regarding timestep definition, previous research has varied—some approaches adopt the start or end time of events, while others abstract timesteps as process stages. In our framework, we define the start time ($T^s_i$) as the timestep to ensure consistent temporal ordering, which is essential for accurate sequential modeling. To address overlapping activities, we explicitly include the duration attribute ($T^d_i$). This allows the model to detect when an event concludes after the subsequent one has begun, effectively capturing simultaneous behavior while maintaining architectural simplicity.

\subsection{Event Label Featurization}
For each event $X_i$ in a sequence $S_j$, we manually map the event (activity) label $\mathcal{A}_i$ into two primary types of semantic attributes: a verb component $\mathcal{A}^v_i$ and a set of descriptive components $\mathcal{A}^{d_k}_i$, where $k$ indexes the descriptors. Each of these attributes is represented as a single word token. The maximum value of $k$ is determined by the structure of the dataset and set to the minimum required for complete coverage.

For example, the event \textit{``Initiate Low Application Check''} is featurized as $\mathcal{A}^v_i = \textit{Check}$, $\mathcal{A}^{d_1}_i = \textit{Low}$, and $\mathcal{A}^{d_2}_i = \textless\texttt{NO\_DESC}\textgreater$, where the special token indicates the absence of a second descriptor. Similarly, \textit{``Check Insurance History''} is featurized as $\mathcal{A}^v_i = \textit{Check}$, $\mathcal{A}^{d_1}_i = \textit{Insurance}$, and $\mathcal{A}^{d_2}_i = \textit{History}$, while \textit{``Check Insurance Payment''} becomes $\mathcal{A}^v_i = \textit{Check}$, $\mathcal{A}^{d_1}_i = \textit{Insurance}$, and $\mathcal{A}^{d_2}_i = \textit{Payment}$.

The verb attributes ($\mathcal{A}^v_i$) capture the core functional action of the event, while the descriptive attributes ($\mathcal{A}^{d_k}_i$) provide semantic context. This decomposition reduces vocabulary sparsity and enhances generalization by allowing the model to isolate and learn shared structures across events. It also supports more efficient tokenization and embedding by controlling input dimensionality.

Our design is informed by principles from semantic role labeling and structured event modeling, which emphasize verb-centric representations as foundational in both natural language processing and process mining \cite{pustejovsky2012natural,weber2008change}. After featurization, the original label $\mathcal{A}_i$ is replaced with $\mathcal{A}'_i$, formed by concatenating $\mathcal{A}^v_i$ and all $\mathcal{A}^{d_k}_i$ using underscores (``\_'') as separators.

\subsection{Pseudo-embedding Universal Attributes}
To embed contextual attribute and temporal information without relying on external models, we introduce two term frequency-inverse document frequency (tf-idf) based pseudo-embedding methods. Algorithm~\ref{alg:pseudo_embedding} constructs a correlation-based vector representation for each event based on co-occurring universal attributes. Algorithm~\ref{alg:pseudo_bin} extends this strategy to duration patterns via a dynamic binning procedure that balances frequency distributions across short and long duration ranges.
\subsubsection{Pseudo-embedding Attribute Correlations} The pseudo-embedding attribute correlations method captures the relevance of universal attribute combinations to each featurized event $X_i$. This is particularly useful when events depend on multiple contextual attributes—e.g., combinations of destination (home, work) and time of day (morning, night) influencing actions like walking or biking.

We treat each sequence $S_j$ as a document and construct a tf-idf matrix. Each document (i.e., sequence) contains "tokens" formed by concatenating child elements of all universal attributes $U_i$ from each event $X_i \in S_j$. For example, if $U^a_i = \textit{Home}$ and $U^b_i = \textit{Morning}$, the resulting token is ``Home\_Morning".

Let $\mathcal{T}_j$ denote the multiset of such tokens for $S_j$. The tf-idf matrix $M_{\text{cor}} \in \mathbb{R}^{|S_j| \times |V|}$ is built using these tokens, where $|V|$ is the total number of unique attribute combinations (i.e., the vocabulary). The value of $M_{\text{cor}}[i, m] = \text{tf-idf}(\mathcal{A}'_i, U^m)$ quantifies the importance of combination $U^m$ for event $\mathcal{A}'_i$ within the sequence.

Each event $\mathcal{A}'_i$ is then represented as a vector $\mathbf{v}_{\text{cor}_i} \in \mathbb{R}^{|V|}$, where each dimension corresponds to one tokenized attribute combination. Token order is not used in this representation. Algorithm~\ref{alg:pseudo_embedding} provides the procedural steps.

\begin{algorithm}[htbp]
\caption{Construction of a tf-idf-based pseudo-embedding matrix that captures contextual relevance of universal attribute combinations for each event $\mathcal{A'}_i$. Each event is represented as a vector $\mathbf{v}_{cor_i}$ reflecting attribute correlation patterns within its sequence.}
\label{alg:pseudo_embedding}
\begin{algorithmic}[1]
\REQUIRE Set of sequences $\{S_j\}$ and set of events $\{X_i\}$ relabeled as $\mathcal{A}'_i$, where $X_i \in S_j$; set of universal attributes $\{U^n_i\}$ for each $X_i$; set of categorical values $\{C^{U^n_i}_k\}$ for each $U^n_i$, denoted as $U^n_i = \{C^{U^n_i}_k\}$.
\ENSURE tf-idf matrix embedding attribute correlations for each $\mathcal{A}'_i$.

\FOR{each sequence $S_j$}
    \FOR{each event $X_i$ in $S_j$}
        \STATE Extract each attribute combination $U^m$ by computing the Cartesian product over $\{C^{U^a_i}_k, C^{U^b_i}_k, \dots, C^{U^n_i}_k\}$.
        \STATE Create term $(\mathcal{A}'_i, U^m)$ from the featurized label and attribute combination.
    \ENDFOR
\ENDFOR
\STATE Construct a corpus of all $(\mathcal{A}'_i, U^m)$ terms across sequences.
\FOR{each sequence $S_j$}
    \STATE Treat $S_j$ as a document.
    \STATE Compute tf-idf scores for all terms $(\mathcal{A}'_i, U^m)$.
    \STATE Construct a tf-idf matrix where rows correspond to $\mathcal{A}'_i$ and columns to unique $U^m$.
\ENDFOR
\RETURN A pseudo-embedding matrix with each $\mathcal{A}'_i$ represented as a vector $\mathbf{v}_{cor_i}$ across all sequences.
\end{algorithmic}
\end{algorithm}

\subsubsection{Pseudo-embedding Duration Bins}
The pseudo-embedding duration bin method captures the temporal distribution of event durations $T^d_i$ across sequences using a dynamic binning strategy, followed by tf-idf embedding.

We define a cut-off threshold $T_{\text{cut}}$, which separates shorter and longer durations. For $T^d_i < T_{\text{cut}}$, each unique duration forms an individual bin $b$, creating fine-grained representation for frequently occurring short durations. For $T^d_i \geq T_{\text{cut}}$, we apply $q$-quantile binning to partition the long-duration values into $q$ equal-sized bins, denoted $b^*_1, \dots, b^*_q$.

Each event $X_i$ is thus assigned to a bin $T^{d_{b'}}_i$, where $b' \in \{b\} \cup \{b^*_1, \dots, b^*_q\}$. We tokenize each event by concatenating the updated activity label $\mathcal{A}'_i$ with its assigned duration bin (e.g., ``Check$\_b^*_3$''), creating a set of tokens $\mathcal{D}_j$ for sequence $S_j$.

A tf-idf matrix $M_{\text{bin}} \in \mathbb{R}^{|S_j| \times |B|}$ is constructed, where $|B|$ is the number of unique bin-labeled tokens across all sequences. Each row corresponds to an event, and each column to a specific $\mathcal{A}'_i$–duration bin pair. Each event vector $\mathbf{v}_{\text{bin}_i} \in \mathbb{R}^{|B|}$ encodes its temporal context within the sequence. Algorithm~\ref{alg:pseudo_bin} provides a step-by-step outline.

\begin{algorithm}[htb!]
\caption{Construction of a pseudo-embedding matrix using dynamically assigned duration bins. Each event $\mathcal{A'}_i$ is represented as a vector $\mathbf{v}_{bin_i}$ based on tf-idf scores of its duration bin, reflecting the frequency and relevance of temporal patterns.}

\label{alg:pseudo_bin}
\begin{algorithmic}[1]
\REQUIRE Set of events $\{X_i\}$ (featurized and relabeled as $\mathcal{A'}_i$); duration values $T^d_i$ for each event; set of sequences $\{S_j\}$.
\ENSURE Pseudo-embedding matrices with each $\mathcal{A'}_i$ represented as a vector $\mathbf{v}_{bin_i}$ across all sequences.

\STATE Initialize cut-off threshold $T_{\text{cut}}$ and number of quantile-based bins $\mathcal{N}_{b*}$.

\REPEAT
    \FOR{each event $\mathcal{A'}_i$}
        \IF{$T^d_i < T_{\text{cut}}$}
            \STATE Assign $T^d_i$ to a unique bin $b$ (fine-grained binning for short durations).
        \ELSE
            \STATE Apply quantile-based binning to $T^d_i$ using $\mathcal{N}_{b*}$ bins, denoted as $b^*$.
            \STATE Adjust $b^*$ to ensure full range coverage and remove duplicate boundaries.
        \ENDIF
        \STATE Assign the final duration bin $T^{d_{b'}}_i \in \{b, b^*\}$ to event $\mathcal{A'}_i$.
    \ENDFOR

    \STATE Calculate bin frequencies $\{f_{b'}\}$ across all events.

    \IF{all $f_{b^*}$ are approximately balanced}
        \STATE BREAK
    \ELSE
        \STATE Adjust $T_{\text{cut}}$ and/or $\mathcal{N}_{b*}$.
    \ENDIF
\UNTIL{stopping condition is met (e.g., sufficient balance in bin frequencies).}

\STATE Extract unique combinations $(\mathcal{A'}_i, T^{d_{b'}}_i)$ and treat each as a term.
\STATE Construct a corpus from all such terms across sequences.

\FOR{each sequence $S_j$}
    \STATE Treat $S_j$ as a document.
    \STATE For each term $(\mathcal{A'}_i, T^{d_{b'}}_i)$ in $S_j$, compute $\text{tf-idf}(\mathcal{A'}_i, T^{d_{b'}}_i)$.
    \STATE Construct a tf-idf matrix with rows as $\mathcal{A'}_i$ and columns as $T^{d_{b'}}_i$.
\ENDFOR

\RETURN Pseudo-embedding matrices where each $\mathcal{A'}_i$ is embedded as vector $\mathbf{v}_{bin_i}$.
\end{algorithmic}
\end{algorithm}
For both pseudo-embedding attribute correlations and duration bin methods, the resulting tf-idf scores are normalized by using Min-Max scaling. Through applying both methods, each event is embedded in a high-dimensional space, where each dimension corresponds to a specific attribute or duration bin, reflecting its contextual significance.

\subsection{Simultaneous Events Vectorization}
\subsubsection{Multidimensional Embedding}
To address the representation of simultaneous events, we propose a multidimensional embedding scheme. Unlike simple multi-hot encodings, this approach accommodates the heterogeneity of event attributes—both categorical and numerical. Each event $X_i$, along with its associated attributes $F'_i = [\mathcal{A}^{v}_i, \mathcal{A}^{d_k}_i, F_i]$, is embedded into a dense vector, as $
\mathbf{v}_i = \text{Embed}(X_i, F'_i)$.

When multiple events occur simultaneously (denoted as $X_{\text{co}}$), their embeddings are concatenated to form a composite representation $\mathbf{v}_{\text{co}} = \text{Concatenate}([\mathbf{v}_1, \mathbf{v}_2, \ldots, \mathbf{v}_n])$

Although summation or averaging could serve as aggregation strategies, concatenation is preferred here to preserve attribute specificity—particularly where numerical features are present only in a subset of events. 

\subsubsection{Time-Difference Flag Augmentation}
To further support the encoding of temporal context, we augment each event vector with a time-difference feature $\Delta T_i = T^s_i - T^s_{i-1}$. For truly simultaneous events, $\Delta T_i$ remains constant across all instances within a given timestep, implicitly indicating co-occurrence.

Overall, this vectorization strategy captures the structural and temporal intricacies of simultaneous events, enabling the model to learn from both attribute-rich encodings and fine-grained temporal signals. This approach may also be integrated with pseudo-embedding mechanisms to enrich the model’s input space.

\section{LSTM HyperModels}
\label{sec:HM}
\subsection{LSTM HyperModels Architectures}
\subsubsection{Base LSTM Model (B-LSTM}
The base LSTM model processes each event \( X_i \) within a sequence \( S_j \) by combining categorical (\( C_{X_i}, C_{S_j} \)) and numerical attributes (\( N_{X_i}, N_{S_j} \)) into feature vectors \( \mathbf{v}_{X_i} \) and \( \mathbf{v}_{S_j} \), defined as $
\mathbf{v}_{X_i} = [C_{X_i}, N_{X_i}]$, and $ \quad \mathbf{v}_{S_j} = [C_{S_j}, N_{S_j}]$. \textbf{F-B-LSTM Variant:} For the time-augmented variant (F-B-LSTM), the numerical attribute vector includes the time difference feature \( \Delta T_i \), enabling the model to consider temporal intervals between events. \textbf{M-B-LSTM Variant:} For the multidimensional embedding variant (M-B-LSTM), the event-level input vector is extended to \( \mathbf{v}^{\text{co}}_{X_i} \), which incorporates learned co-occurrence embeddings across simultaneous events and sub-status annotations.

For categorical attributes with missing descriptors, we use the placeholder ``\(\langle \text{NO\_DESC} \rangle\)” encoded as \(-1\), and during training, these values are masked. Numerical attributes with missing values are replaced by the median (\( \widetilde{N_{X_i}} \)) to mitigate data skewness.

The architecture starts with the event input \( \mathbf{v}_{X_i} \), which is processed through a stack of LSTM layers \( l^e_\rho \) (where \( \rho \) denotes the layer number).  The final LSTM layer \( l^e_\mathcal{P} \) outputs a sequence representation \( \mathbf{v}_{S'_j} \), which is concatenated with the sequence-level input \( \mathbf{v}_{S_j} \) to form a combined representation \( Z \): $ Z = \text{Concatenate}(\mathbf{v}_{S'_j}, \mathbf{v}_{S_j})$.

This combined representation is subsequently passed through fully connected layers, with the final output layer applying the Softmax function for categorical classification. The B-LSTM model effectively captures both event and sequence level dependencies, enhancing predictive accuracy.

\subsubsection{Pseudo-Embedding LSTM}
This study introduces three variants of the Pseudo-Embedding LSTM: the Pseudo-Embedding Duration LSTM (D-LSTM), its time-augmented variant (F-D-LSTM) for handling simultaneous events, and the Duration-Correlation Pseudo-Embedding LSTM (DC-LSTM). All models use the same configuration for the input vectors \( \mathbf{v}_{X_i} \) and \( \mathbf{v}_{S_j} \) as in the base LSTM (B-LSTM), with F-D-LSTM inheriting the time-difference augmentation from F-B-LSTM. Each event input \( \mathbf{v}_{X_i} \) is processed through a stack of LSTM layers \( l^e_\rho \), and the output \( \mathbf{v}'_{X_i} \) of the final LSTM layer \( l^e_\mathcal{P} \) retains the shape of the input.

In D-LSTM and F-D-LSTM, an additional input vector \( \mathbf{v}_{\text{bin}_i} \) is processed through a separate stack of LSTM layers, producing the output \( \mathbf{v}_{\text{bin}'_i} \). A concatenated vector \( \Psi \) is formed as $\Psi = \text{Concatenate}(\mathbf{v}_{X'_i}, \mathbf{v}_{\text{bin}'_i})$.

In the DC-LSTM, a third input vector \( \mathbf{v}_{\text{cor}_i} \) is introduced and processed through its own LSTM layers. The concatenated vector \( \Psi \) is then updated to include the output of this third vector $\Psi = \text{Concatenate}(\mathbf{v}_{X'_i}, \mathbf{v}_{\text{bin}'_i}, \mathbf{v}_{\text{cor}'_i})$.

Finally, the combined vector \( \Psi \) is passed through additional LSTM layers, producing the sequence output \( \mathbf{v}_{S'_j} \). This output is combined with \( \mathbf{v}_{S_j} \) using the same procedure as in B-LSTM to form the final representation \( Z \), which is then processed through dense layers to yield the final output.

\subsubsection{Textual Embedding LSTM (T-LSTM)}
The T-LSTM architecture extends the baseline by incorporating NLP-inspired embeddings for textual activity descriptors. For each event $X_i$, the featurized components $\mathcal{A}^v_i$ and $\mathcal{A}^{d_k}_i$ are treated as textual tokens and converted into vectors $\mathbf{v}^v_i$ and $\mathbf{v}^{d_k}_i$. These are passed through embedding layers to obtain $\mathbf{E}^v_i = \text{Embed}(\mathbf{v}^v_i)$ and $\mathbf{E}^{d_k}_i = \text{Embed}(\mathbf{v}^{d_k}_i)$.

The resulting embeddings are concatenated to form a unified representation $ \mathbf{E}^{(v, d_k)}_i = \text{Concatenate}(\mathbf{E}^v_i, \mathbf{E}^{d_k}_i)$, which is then processed through a dedicated LSTM stack, producing output $\mathbf{v}^{(v, d_k)}_i$.

This embedding output is integrated with other inputs, depending on the model variant used (B-LSTM, D-LSTM, or DC-LSTM). The resulting concatenated vector $\Psi$ takes one of the following forms:

\begin{flalign}
&& \Psi &= \text{Concatenate}(\mathbf{v}^{(v, d_k)}_i, \mathbf{v}_{X'_i}), &\\
&& \Psi &= \text{Concatenate}(\mathbf{v}^{(v, d_k)}_i, \mathbf{v}_{X'_i}, \mathbf{v}_{\text{bin}'_i}), &\\
&& \Psi &= \text{Concatenate}(\mathbf{v}^{(v, d_k)}_i, \mathbf{v}_{X'_i}, \mathbf{v}_{\text{bin}'_i}, \mathbf{v}_{\text{cor}'_i}) \footnotemark. &
\end{flalign} \footnotetext{For demonstration purposes, we use this most comprehensive form in our implementation.}

The combined vector $\Psi$ is then passed through additional LSTM layers to generate the sequence-level output $\mathbf{v}_{S'_j}$, which is subsequently merged with $\mathbf{v}_{S_j}$ as in the B-LSTM. The final representation $Z$ is processed through dense layers to produce the output prediction.

\subsection{Hyperparameter Selection and Justification}
In this study, LSTM models are developed to predict the outcomes of business process instances using proposed encoding and embedding strategies. To enhance model adaptability across diverse event logs, we employ a dynamic LSTM HyperModel approach, where hyperparameters are automatically tuned based on input characteristics. This approach facilitates the creation of multiple benchmark pipelines for sequence classification. Hyperparameter selection is guided by both theoretical foundations and empirical findings in deep learning. Table \ref{tab:hyperparameters} presents the tuning ranges, and the rationale for key hyperparameters is provided below.

\begin{table}[htbp!]
\begin{threeparttable}
\centering
\caption{Hyperparameters and Their Tuning Ranges/Types}
\label{tab:hyperparameters}
\begin{tabularx}{\linewidth}{lX}
\hline
\textbf{Hyper-P}                          & \textbf{Range/Type}                            \\\toprule
\multicolumn{2}{l}{\textit{\textbf{LSTM Layers}}}
\\
$\#$ & 1$\sim$3                                    \\ 
Units                    & 16$\sim$512 (step 16)                       \\
L2$^\dagger$ & $1\times10^{-5}\sim1\times10^{-2}$  \\ 
Batch Norm &Y/N; MM$^\dagger$:0.01$\sim$0.999; eps$^\dagger$: {\scriptsize$1 \times 10^{-5}\sim1 \times 10^{-2}$}           \\ 
Dropouts                 &Y/N; rates :0.2$\sim$0.7                                \\ 
\midrule
\multicolumn{2}{l}{\textit{\textbf{Dense Layers}}}
\\
$\#$     & 1$\sim$3                                    \\
Units                 & 16$\sim$256 (step 16)                       \\ 
L2$^\dagger$       & $1 \times 10^{-5}\sim1 \times 10^{-2}$  \\ 
Dropouts                 &Y/N; rates :0.2$\sim$0.7                                \\
Activation           & ReLU, Tanh, Softmax, Leaky\_ReLU ($\alpha:       0.01\sim0.3 $)                            \\ 
\midrule
\multicolumn{2}{l}{\textit{\textbf{Learning Rate (LR) Scheduler and Optimizer}}}
\\
LR         & \scriptsize{Exponential, Inverse Time, Piecewise\_Constant, Polynomial }    \\
Initial LR        & $1 \times 10^{-4}\sim1 \times 10^{-2}$  \\ 
Optimizer                    & Adam, SGD, RMSprop                        \\ 
Adam               & $\beta_1:0.85\sim0.99                          $; $\beta_2 :0.99\sim0.999$                           \\ 
SGD               & MM$^\dagger$:0.0$\sim$0.9                                \\ 
RMSprop & {\scriptsize$\alpha: 0.9\sim0.999$; MM$^\dagger$: 0.01$\sim$0.9; eps:$1\times 10^{-6} \sim 1\times 10^{-10}$}\\
\midrule
Embedding$^\ddagger$  & 10$\sim$250 (step 10)
\\ Batch Size & 16, 31, 64, 128
\\\bottomrule
\end{tabularx}
    \begin{tablenotes}
	\small {
	\item{ $\dagger$ L2: l2 regularization; MM: momentum; eps:epsilon}
	\item $\ddagger$ Only for the verb and description of activity in T-LSTM}
    \end{tablenotes}
\end{threeparttable}
\vspace{-10pt}
\end{table}

First, the depth and width of LSTM networks impact their ability to capture long-range dependencies in event sequences. Deeper architectures improve representational capacity, but excessive depth can lead to vanishing gradient issues, which compromise training stability. The search range is designed to balance model expressiveness with computational feasibility, following best practices in recurrent network design \cite{graves2012supervised, tax2017predictive}. To prevent overfitting, L2 regularization is applied within the range of $1 \times 10^{-5}$ to $1 \times 10^{-2}$, as it effectively controls model complexity \cite{cheng2017exploration}. Dropout is optional with rates from 0.2 to 0.7, promoting generalization by randomly deactivating neurons during training \cite{gal2016theoretically}. Additionally, batch normalization (flag: Y/N) stabilizes training, with momentum tuned between 0.01 and 0.999 (step 0.1) and epsilon in the range of $1 \times 10^{-5}$ to $1 \times 10^{-2}$ to maintain numerical stability in deep LSTM networks \cite{hinton2012neural, ioffe2015batch}.

Second, dense layers following the LSTM layers refine the features extracted by the LSTM, enhancing predictive performance \cite{glorot2011deep, nguyen2021time}. Shallow architectures (1–2 layers) map LSTM outputs to class probabilities, reducing overfitting \cite{montavon2018methods}. Deeper architectures (3 layers) allow hierarchical feature recombination, capturing more complex decision boundaries \cite{bengio2013deep}. The number of units is selected to balance computational efficiency and representational power, sufficient to handle simple threshold-based decisions while accommodating interactions among multiple event attributes \cite{raghu2017expressive, teinemaa2019outcome}. Activation functions (ReLU, Tanh, Softmax, and Leaky ReLU) are chosen for their ability to introduce nonlinearity and enhance expressiveness, which is essential for capturing event dependencies in PBPM \cite{camargo2019learning}.

Third, the learning rate determines the step size during optimization and affects convergence speed and stability \cite{sutskever2013importance}. We evaluate four decay strategies—Exponential, Inverse Time, Piecewise Constant, and Polynomial—to address different learning dynamics and mitigate overfitting, facilitating faster convergence \cite{yu2020llr}. Business process event logs exhibit varying temporal scales (e.g., short- vs. long-running cases), requiring adaptive decay schedules for optimal training \cite{cahuantzi2023comparison, devlin2019bert}. The choice of optimization algorithm impacts gradient updates and model generalization. We consider Adam, SGD, and RMSprop, as each has distinct advantages for recurrent architectures \cite{kingma2014adam}. Adam’s adaptive moment estimation stabilizes training in sequence modeling tasks \cite{kingma2014adam}, while SGD with momentum ensures robust convergence across both convex and non-convex landscapes \cite{hinton2012neural}. RMSprop, by contrast, improves generalization in non-stationary environments, particularly for long-sequence dependencies \cite{rospawan2025two}.

Finally, the embedding dimension range (10–250) is chosen to align with process log vocabulary characteristics, balancing model expressiveness and computational efficiency \cite{pennington2014glove, mikolov2013efficient}. Batch size impacts training stability and convergence dynamics. Smaller batches introduce gradient noise, helping to escape local minima, while larger batches offer smoother gradients but may converge to sharp minima. The tuning range follows deep learning heuristics to optimize convergence speed and generalization performance \cite{sewak2021lstm}.

\section{Experiment}
\label{sec:EP}
\subsection{Datasets}
This study utilizes four datasets for evaluating model performance: the synthetic *Patients* dataset, and three real-world variants of the BPIC12 dataset—*BPIC12*, *BPIC12-A*, and *BPIC12-O* \cite{BPIC12set}.

\textbf{Patients} is a synthetic healthcare dataset containing 2,140 sequences, each representing a patient's interaction with the healthcare system. It includes both event and sequence level attributes: 3 numerical and 1 categorical at the sequence level, and 3 numerical and up to 3 categorical attributes at the event level (including 1 universal categorical attribute). Each sequence is assigned one of five possible outcomes, with a severe class imbalance: the most common class accounts for 40.74\% of sequences, and the rarest just 1.12\%, leading to an imbalance ratio of approximately 36:1. This attribute-rich and imbalanced structure makes the dataset a suitable benchmark for testing the robustness of LSTM variants, especially those incorporating pseudo-embeddings or correlation-aware modules.

\textbf{BPIC12}, \textbf{BPIC12-A}, and \textbf{BPIC12-O} are derived from real-world loan and overdraft application processes in a multinational financial institution. Each sequence concludes with one of three outcomes: \emph{accepted (approved)}, \emph{declined}, or \emph{canceled}. These datasets were curated to ensure balanced class distributions—*BPIC12-O* includes 802 sequences per class, while *BPIC12* and *BPIC12-A* include 2224 per class. Attribute-wise, they share a relatively simple structure: 1 numerical attribute at the sequence level and 2 universal categorical attributes at the event level. However, they frequently contain multiple events with identical timestamps, making them especially valuable for evaluating temporal augmentation strategies such as those applied in F/M-B-LSTM and F-D-LSTM.

As summarized in Table~\ref{tab:data}, these datasets vary significantly in terms of sequence length, number of cases, and attribute complexity. The *Patients* dataset features shorter but more heterogeneous sequences and richer event-level attributes, which are well-suited to models like DC-LSTM. In contrast, the BPIC12 variants contain simpler sequences but pose challenges related to temporal simultaneity. Sequence lengths range from as few as 3 events to as many as 77, introducing diverse temporal dynamics that can affect model behavior. This diversity across datasets supports a comprehensive evaluation of LSTM-based architectures across varying levels of structural complexity and temporal challenges.

\begin{table}[htb]
\centering
\begin{threeparttable}
\setlength{\tabcolsep}{1.4pt}
\caption{Statistics of the datasets used in the experiments}
\label{tab:data}
\notesize
\begin{tabularx}{\linewidth}{lllllllllllllX}
\hline
data & \#S  & max  & min  & median & \#Attr & \#Attr & Size & \#Attr & \#Attr & Size & \# Outcome\\
set&		&length &length &length &(E,N) & (E,C)& (E,C) & (S,N) &(S,C) &(S,C) &  \\
\toprule
    BPI12 & 6672 & 77   & 12   & 18   & 2    & 0    & -    & 1    & 0    & -    & 3 \\
    BPI12O & 2406 & 30   & 4    & 5    & 2    & 0    & -    & 1    & 0    & -    & 3 \\
    BPI12A & 6672 & 7    & 3    & 6    & 2    & 0    & -    & 1    & 0    & -    & 3 \\
    Patients & 2140 & 9    & 4    & 7    & 3    & 3    & [10,3,3] & 3    & 1    & 2    & 5 \\
\bottomrule
\end{tabularx}
    \begin{tablenotes}[flushleft]
	\scriptsize
\item S: sequence; (E,C): Event level categorical attributes;(E,N): Event level numerical attributes; (S,C): Sequence level categorical attributes;(S,N): Sequence level numerical attributes 
    \end{tablenotes}
\end{threeparttable}
\vspace{-20pt}
\end{table}

\subsection{Attribute Processing}
All datasets include recorded start and completion times for each event. In our experiments, the start time of each event was adopted as the reference time step for sequence modeling. Additionally, we derived a new duration attribute by computing the difference between the start and end times for each event.

To ensure consistency in semantic representation, activity labels were manually featurized across all datasets following our proposed method, as detailed in Table~\ref{tab:featurization}.\footnote{Activity labels in the BPIC12W dataset were originally in German and were translated into English prior to featurization.}
\begin{table}[htbp]
\scriptsize
  \caption{Event Label Featurization}
    \label{tab:featurization}
  \centering
    \begin{tabularx}{\linewidth}{lXX}
    \toprule
    \textbf{Patience Dataset}&&\\
    \midrule
    \textbf{Activity Label} & \textbf{Verb} & \textbf{Description} \\ 
    Registration & register & \textless NO\_DESC\textgreater \\ 
    Basic Check & check & basic \\ 
    Initiate Low Application Check & check & low \\ 
    Check Insurance History & check & insurance \\ 
    Check Medical History & check & medical \\ 
    Send Notification & note & \textless NO\_DESC\textgreater \\ 
    Archive & archive & \textless NO\_DESC\textgreater \\ 
    Receive Questionnaire & question & \textless NO\_DESC\textgreater \\ 
    Initiate High Application Check & check & high \\ 
    Check Hospital Records & check & hospital \\ \bottomrule
    \textbf{BPIC12 Dataset}&&\\
    \midrule
    \textbf{Activity Label} & \textbf{Verb} & \textbf{Description} \\ 
    ACCEPTED & accept & \textless NO\_DESC\textgreater \\
ACTIVATED & activate & \textless NO\_DESC\textgreater \\
APPROVED & approve & \textless NO\_DESC\textgreater \\
FINALIZED & finalize & \textless NO\_DESC\textgreater \\
PARTLYSUBMITTED & submit & partial \\
PREACCEPTED & accept & pre \\
REGISTERED & register & \textless NO\_DESC\textgreater \\
SUBMITTED & submit & \textless NO\_DESC\textgreater\\
CREATED & create & \textless NO\_DESC\textgreater \\
SELECTED & select & \textless NO\_DESC\textgreater\\
SENT & send & \textless NO\_DESC\textgreater\\
SENT\_BACK & return & \textless NO\_DESC\textgreater \\
CANCELLED & cancel & \textless NO\_DESC\textgreater \\
COMPLETE & complete & \textless NO\_DESC\textgreater\\
QUOTE & quote & \textless NO\_DESC\textgreater\\
HANDLE & handle & \textless NO\_DESC\textgreater\\
FOLLOW & follow & \textless NO\_DESC\textgreater\\
ASSESS & assess & \textless NO\_DESC\textgreater\\
 \bottomrule
    \end{tabularx}%
    \vspace{-18pt}
\end{table}%

To compare the performances of B-LSTM, D-LSTM, DC-LSTM, and T-LSTM on the Patients dataset, we applied our proposed pseudo-embedding attribute methods. For the duration-based embedding, event durations were first converted from seconds to minutes and then rounded up to the nearest integer. These rounded durations were then segmented into 24 bins: durations under 5 minutes were placed into individual bins per unique value, while durations greater than or equal to 5 minutes were grouped into quantile-based bins. This binning strategy ensured a more balanced distribution of values for downstream embedding. The resulting bins were embedded using a fixed embedding layer of size 24. Additionally, since the Patients dataset contains only one universal attribute, a dummy attribute was added to enable the processing of pseudo-embedding correlations in DC-LSTM.

For the BPIC12 dataset, durations were discretized into two bins: one representing events with zero duration and the other for non-zero durations. To evaluate different encoding and embedding strategies under simultaneous event conditions, we applied F-B-LSTM and M-B-LSTM to the BPIC12 and BPIC12-A/O datasets. Notably, F-D-LSTM was only applied to BPIC12, as it is the only variant where event durations exhibit meaningful variation.

\subsection{LSTM Hyperparameters Searching}The proposed LSTM HyperModels underwent hyperparameter optimization using the Hyperband algorithm, selected for its effectiveness in balancing exploration and exploitation across large search spaces. Hyperband was configured to maximize validation accuracy for balanced datasets and validation weighted F1-score for imbalanced datasets, with a maximum of 300 epochs and a reduction factor of 3. Each model was trained using an 80/20 train-validation split, and early stopping was employed to prevent overfitting.

After the tuning process, optimal hyperparameters were extracted from the best-performing trial. Final evaluation results were obtained by either retrieving the best model directly or rebuilding it using the selected hyperparameters. As the task involves multiclass outcome prediction for each sequence, we report accuracy for balanced datasets and weighted F1-score for imbalanced datasets to ensure a fair and comprehensive performance evaluation. 

\subsection{Computational Overhead of Hierarchical Modeling}
To assess the complexity of the two-level hierarchical framework, we report the training time for all models across datasets in Table~\ref{tab:cost}. Experiments were conducted on a system with an Intel(R) Core(TM) i9-8950HK CPU (6 cores, 12 threads) and 32 GB RAM, without GPU acceleration. As expected, models incorporating richer representations—such as DC-LSTM and F-D-LSTM—require longer training times (up to 7.5 hours per dataset), while simpler baselines like B-LSTM complete in under 2 hours. This variation reflects the increased computational cost introduced by the hierarchical architecture and hyperparameter tuning process. While the overhead is non-trivial, it remains acceptable for offline predictive business process monitoring, where accuracy and generalizability are key. Future work may explore model compression techniques to improve efficiency in real-time or resource-constrained environments.
\begin{table}[htb]
\centering
\setlength{\tabcolsep}{1.3pt}
\caption{Computational Cost}
\label{tab:cost}
\notesize
\begin{tabularx}{\linewidth}{cccccccX}
\hline
\multicolumn{4}{c}{Patients Dataset}& \multicolumn{4}{c}{BPIC12 Datasets}\\
\cmidrule[0.3pt](lr){1-4}\cmidrule[0.3pt](lr){5-8}
 B-LSTM & D-LSTM & DC-LSTM & T-LSTM &      & M-B-LSTM & F-B-LSTM & F-D-LSTM \\
\cmidrule[0.3pt](lr){1-4}\cmidrule[0.3pt](lr){5-8}
    1h30m43s & 4h47m18s & 6h05m20s & 5h10m17s & 12 & 5h51m27s & 4h16m44s & 7h38m52s \\
         &      &      &      & 12O & 2h23m14s & 1h57m25s &  \\
         &      &      &      & 12A & 3h34m21s & 1h32m34s &  \\
\bottomrule
\end{tabularx}
\vspace{-20pt}
\end{table}

\section{Results} 
\label{sec:RE}
\subsection{Architectures and Hyperparameters}
Table~\ref{tab:HyperP} summarizes the architectures and optimized hyperparameters of the proposed LSTM HyperModels across all datasets. Each configuration was tailored to align with the specific encoding and embedding strategies employed, aiming to maximize performance. Variations in layer depth, dropout rates, and other hyperparameters reflect dataset-specific trade-offs between model capacity, regularization, and learning stability.

\begin{table}[htbp]
  \centering
  \begin{threeparttable}
  \caption{Hyperparameters and Architectures of LSTM HyperModels}
  \label{tab:HyperP}
  \scriptsize
\noindent
\begin{tabularx}{\linewidth}{@{}l@{\hspace{0pt}}c@{\hspace{0pt}}c@{\hspace{1pt}}c@{\hspace{1pt}}c@{\hspace{1pt}}c@{\hspace{1pt}}c@{\hspace{0.5pt}}c@{\hspace{0.5pt}}c@{\hspace{1pt}}c@{\hspace{1pt}}c@{\hspace{1pt}}c@{\hspace{1pt}}c}
\toprule
    M     & B     & L(U)  & L(D)& L(Bm) & L(Be) & L(l2) & D(U)  & D(D)  & D(l2) &  D(A)    & Opt     & LR \\
\cmidrule[0.8pt](r){1-2}    
\cmidrule[0.8pt](r){3-7}
\cmidrule[0.8pt](lr){8-11}
\midrule
    B     & 32    & 160   & 0.4914 & 0.81 &3.345e-4 & 1.956e-4 & 144   & 0.4581 & 2.017e-4 &  ReLU    & (Adam     & Exp \\
          &       & 48    & 0.3156 &      &      & 4.433e-3 &       &       &       & 0.93  & 0.992) & 2.718e-3\\
\midrule
  D     & 16    & 256   & 0.2088 & 0.61  & 6.736e-4 & 1.265e-3 & 192   & 0.4401 & 2.857e-3 & (l\_rl & rms   & P-C \\
\cmidrule(r){3-7}
          & (d)     & 256   & 0.4085 &       &       & 2.99e-3 & 256   & 0.2622 & 9.855e-5 & 0.1997) &       & 5.48e-4 \\
          &       & 64    & 0.3875 & 0.11  & 2.592e-5 & 9.411e-3 &       &       &       & ReLU  &       &  \\
\cmidrule(r){3-7}
          & (c)     & 128   & 0.3635 &       &       & 1.121e-4 &       &       &       &       &       &  \\
          &       & 96    & 0.4356 & 0.21  & 3.468e-4 & 1.14e-4 &       &       &       &       &       &  \\
\midrule          
    C    & 64    & 160   & 0.4875 & 0.11  & 9.891e-5 & 2.777e-5 & 96    & 0.4566 & 2.702e-4 & (l\_rl & (Adam & I-T \\
\cmidrule(r){3-7}
          & (d)     & 32    & 0.3305 &       &       & 3.63e-3 & 160   & 0.1334 & 2.509e-5 & 0.2628) & 0.88  & 8.904e-4 \\
          &       & 64    & 0.4504 & 0.41  & 8.259e-3 & 5.129e-4 &       &       &       & (l\_rl & 0.994) &  \\
\cmidrule(r){3-7}
          & (cr)    & 64    & 0.2501 &       &       & 5.959e-3 &       &       &       & 0.1600) &       &  \\
\cmidrule(r){3-7}
          & (c)     & 32    & 0.2927 &       &       & 1.674e-4 &       &       &       &       &       &  \\
\midrule
    T     & 32    & 32    & 0.3215 & 0.61  & 6.057e-5 & 7.91e-5 & 96    & 0.4837 & 9.832e-4 & tanh  & (Adam & Poly \\
\cmidrule(r){3-7}
          & (e)     & 192   & 0.2307 & 0.01  & 3.343e-4 & 7.738e-3 &       &       &       &       & 0.99  & 3.611e-3 \\
          & \tiny{$10^\dagger$}    & 128   & 0.3076 &       &       & 6.582e-3 &       &       &       &       & 0.996) &  \\
          & \tiny{$40^\dagger$}    & 64    & 0.2456 &       &       & 1.522e-4 &       &       &       &       &       &  \\
\cmidrule(r){3-7}
          & (d)     & 64    & 0.4809 &       &       & 1.248e-4 &       &       &       &       &       &  \\
          &       & 96    & 0.2327 & 0.81  & 8.913e-3 & 6.854e-5 &       &       &       &       &       &  \\
          &       & 256   & 0.2903 &       &       & 1.375e-3 &       &       &       &       &       &  \\
\cmidrule(r){3-7}
          & (cr)    & 224   & 0.408 & 0.11  & 1.464e-4 & 1.413e-4 &       &       &       &       &       &  \\
\cmidrule(r){3-7}
          & (c)     & 96    & 0.3111 &       &       & 1.632e-5 &       &       &       &       &       &  \\
          &       & 96    & 0.2668 & 0.51  & 5.048e-4 & 2.825e-5 &       &       &       &       &       &  \\
\midrule
              M     & 128   & 80    & 0.2381 & 0.01  & 1.033e-5 & 4.713e-5 & 16    & 0.5898 & 4.719e-3 & Soft  & rms   & Exp \\
    (a)   &       &       &       &       &       &       & 16    & 0.1   & 1.002e-5 & ReLU  &       & 6.425e-4 \\
 \cmidrule(r){2-13}
    (o)   & 32    & 80    & 0.4228 & 0.71  & 1.14e-4 & 2.021e-3 & 112   & 0.4477 & 5.367e-3 & soft  & rms   & Exp \\
 \cmidrule(r){2-12}
    (w)   & 64    & 32    & 0.2082 & 0.71  & 1.271e-3 & 1.66e-5 & 208   & 0.1163 & 6.795e-4 & tanh  & (Adam & 7.131e-3  \\
 \cmidrule(r){13-13}
          &       & 96    & 0.3024 &       &       & 1.899e-5 &       &       &0.96     & 0.997)  & Poly & 9.441e-3 \\
\midrule
    F     & 16    & 64    & 0.4031 & 0.61  & 1.041e-4 & 2.274e-5 & 144   & 0.2432 & 1.772e-4 & tanh  & rms   & poly \\
    (a)   &       &       &       &       &       &       & 176   & 0.2433 & 1.722e-3 & tanh  &   & 1.154e-3 \\
\cmidrule(r){2-13}
    (o)   & 16    & 80    & 0.4763 & 0.01  & 1.004e-5 & 8.741e-5 & 144   & 0.3808 & 7.703e-5 & ReLU  & (Adam & Poly \\
          &       & 160   & 0.3397 &       &       & 1.767e-3 & 48    & 0.3603 & 2.354e-4 & ReLU  & 0.88  & 4.306e-3 \\          
\cmidrule(r){2-11}
\cmidrule(r){13-13}
    (w)   & 16    & 160    & 0.3449 & 0.01  & 1.022e-5 & 1.351e-5 & 80    & 0.1946 & 1.868e-3 & (l\_rl & 0.993) & Exp \\
\cmidrule{12-12}
          &       & 224    & 0.2   &       &       & 1.011e-5 & 16    & 0.1   & 1.001e-5 & 0.01) & rms   & 7.933e-3 \\
          &       &       &       &       &       &       & 16    & 0.1123 & 1.012e-4 & ReLU  &       &  \\
\midrule
    U     & 128   & 224   & 0.3647 &       &       & 1.41e-3 & 224   & 0.5757 & 1.098e-4 & tahn  & (Adam & Poly \\
\cmidrule(r){3-7}
   (w)    & (d)     & 32    & 0.3106 & 0.81  & 2.403e-3 & 2.103e-5 &       &       &       &       & 0.91  &  \\
\cmidrule(r){3-7}
          & (c)     & 160   & 0.3911 & 0.51  & 8.522e-5 & 1.567e-5 &       &       &       &       & 0.991) &  \\  
\bottomrule
    \end{tabularx}%
    \begin{tablenotes}[flushleft]
    \scriptsize
	\item[$\bullet$] M: Model; B: B-LSTM; D: D-LSTM; C:DC-LSTM; T:T-LSTM; M:M-B-LSTM; F:F-B-LSTM; U: F-D-LSTM; (a)/(o)/(w):BPI12-A/O/BPI12 datasets.
	\item[$\bullet$]  B: Batch size; L(U)/D(U): Hidden Unis of L(LSTM)/D(Dense) layers; L(D)/D(D): Dropout rates; L(l2)/D(l2): L2 regularization; L(Be): Batch normalization epsilon; L(Bm): Batch normalization momentum; D(A): Activation functions; Opt:Optimizer; LR: Learning rate scheduler(top) and learning rate (bottom); Empty cell in (Bm)/(Be)/(D): No batch normalization or dropout applied. 
	\item[$\bullet$](d): Pseudo-Embedding duration input layer; (cr): Pseudo-Embedding correlation input layer; (c): Concatenation LSTM layer; (e): Embedding layer; $\dagger$: Verb (top) and description (bottom) embedding dimensions.
	\item[$\bullet$] l\_rl: Leaky\_ReLU; soft:softmax; Exp: Exponential; I-T: Inverse Time; P-C: Piecewise\_Constant; Poly: Polynomial.
    \end{tablenotes}
\end{threeparttable}
\vspace{-18pt}
\end{table}%

The B-LSTM model employs dual LSTM layers for event inputs, reflecting a design focused on capturing intricate sequential dependencies from multiple perspectives. To mitigate overfitting risks associated with high-dimensional inputs, the model applies substantial dropout and L2 regularization. The Adam optimizer, combined with exponential learning rate decay, facilitates stable and adaptive convergence. A moderately sized dense layer balances expressiveness and regularization, retaining critical patterns while keeping complexity manageable.

The D-LSTM model adopts a modular structure, comprising an LSTM layer for event inputs, two duration pseudo-embedding layers, and two fusion layers for combining representations. This separation of temporal information suggests clearer modeling of input-output dependencies. RMSprop is selected for its ability to adjust to non-stationary input distributions via parameter-specific updates, aligning well with the model's heterogeneity. A piecewise constant learning rate schedule assists convergence by reducing the learning rate at defined intervals. The denser final layer compared to B-LSTM supports the broader representational demands of its integrated inputs.

The DC-LSTM model features a streamlined structure, with a single LSTM layer dedicated to correlation embedding, likely simplified by the presence of dummy variables that reduce feature interaction complexity. Its dense layers employ moderate unit sizes and varied dropout rates to balance capacity and regularization. The Adam optimizer, in tandem with inverse time learning rate scheduling, provides adaptive control across training, enabling the model to refine performance gradually. This configuration effectively integrates diverse pseudo-embedded features while maintaining structural efficiency.

The T-LSTM model incorporates verb and description embeddings through multiple LSTM layers, applied to concatenated verb-decoded vectors alongside duration and correlation pseudo-embeddings. This architecture addresses the multi-faceted nature of event data. The use of the Adam optimizer with polynomial learning rate scheduling provides the flexibility and control needed for deeper networks. Despite its complexity, the model concludes with a low-unit dense layer, indicating that earlier layers have sufficiently enriched the feature representations. Compared to DC-LSTM, the design reduces dense layer complexity while preserving expressiveness.

Among models designed to handle simultaneous events, the LSTM hypermodels exhibit several recurring patterns. One notable observation is that M-B-LSTM models tend to adopt simpler LSTM configurations with fewer layers. This suggests that their multidimensional encoding strategy effectively captures simultaneity without requiring deep architectures. In contrast, F-B-LSTM models typically require deeper networks or larger hidden units to process time-difference flag encodings, indicating a need for increased model capacity to interpret temporal signals. Another distinction lies in the design of dense layers. M-B-LSTM models often utilize fewer dense layers but vary activation functions—such as softmax, tanh, and ReLU—to balance classification accuracy and intermediate feature abstraction. F-B-LSTM models, by comparison, use additional dense layers with Tanh and ReLU activations, highlighting a greater emphasis on nonlinear transformations and enriched feature representation. This difference implies that each encoding strategy produces feature interactions that behave differently when aligned with sequence-level inputs. Moreover, both model types leverage adaptive optimization strategies—typically combinations of RMSprop or Adam—paired with exponential or polynomial learning rate schedules. This consistency across models reflects a shared need to manage heterogeneous input dynamics and ensure stable convergence. Finally, when comparing D-LSTM and F-D-LSTM, the latter includes simultaneous event encoding yet maintains a simpler architecture: each input stream passes through only one LSTM layer. This implies that the time-difference flag encoding may lead to clearer separability between events, thereby reducing the necessity for deeper recurrent structures.

\subsection{Performance Evaluation}
\subsubsection{Sequential Outcome Prediction Results}
Table \ref{tab:confusion} presents the classification reports for each model on the patients dataset, detailing precision, recall, and F1-score metrics across individual outcome classes. Several key insights emerged from fine-tuning various LSTM HyperModels on patients sequences. 

\begin{table}[!ht]
	\begin{threeparttable}
		\setlength{\tabcolsep}{0.5pt}
		\centering
		\caption{Classification Report of LSTM Models for Patients Dataset}
		\label{tab:confusion}%
		\notesize
		\begin{tabularx}{\linewidth}{cccccccccccccX}
			\toprule
			C & \multicolumn{3}{>{\hsize=\dimexpr3\hsize}c}{\textbf{B-LSTM}}
			& \multicolumn{3}{>{\hsize=\dimexpr3\hsize}c}{\textbf{D-LSTM}} 
			& \multicolumn{3}{>{\hsize=\dimexpr3\hsize}c}{\textbf{DC-LSTM}}
			& \multicolumn{3}{>{\hsize=\dimexpr3\hsize}c}{\textbf{T-LSTM}} 
			& S \\ \cmidrule[0.8pt](lr){2-4}\cmidrule[0.8pt](lr){5-7}\cmidrule[0.8pt](lr){8-10} \cmidrule[0.8pt](lr){11-13}  \midrule 
			0 & \cellcolor[rgb]{ .792,  .929,  .984}1      & \cellcolor[rgb]{ .969,  .78,  .675}1      & \cellcolor[rgb]{ .8,  1,  .6}1      & 							\cellcolor[rgb]{ .792,  .929,  .984}1      & \cellcolor[rgb]{ .969,  .78,  .675}1      & \cellcolor[rgb]{ .8,  1,  .6}1      &							\cellcolor[rgb]{ .792,  .929,  .984}1      & \cellcolor[rgb]{ .969,  .78,  .675}1      & \cellcolor[rgb]{ .8,  1,  .6}1      & 							\cellcolor[rgb]{ .792,  .929,  .984}1      & \cellcolor[rgb]{ .969,  .78,  .675}1      & \cellcolor[rgb]{ .8,  1,  .6}1      & 92 \\                                 

			1 & \cellcolor[rgb]{ .792,  .929,  .984}0.8095 & \cellcolor[rgb]{ .969,  .78,  .675}0.9770 & \cellcolor[rgb]{ .8,  1,  .6}0.8854 & 
				\cellcolor[rgb]{ .792,  .929,  .984}0.8047 & \cellcolor[rgb]{ .969,  .78,  .675}0.9943 & \cellcolor[rgb]{ .8,  1,  .6}0.8895 & 
				\cellcolor[rgb]{ .792,  .929,  .984}0.7803 & \cellcolor[rgb]{ .969,  .78,  .675}1      & \cellcolor[rgb]{ .8,  1,  .6}0.8766 &
				\cellcolor[rgb]{ .792,  .929,  .984}0.7758 & \cellcolor[rgb]{ .969,  .78,  .675}0.9943 & \cellcolor[rgb]{ .8,  1,  .6}0.8715 &174 \\                      		
			2 & \cellcolor[rgb]{ .792,  .929,  .984}0.7143 & \cellcolor[rgb]{ .969,  .78,  .675}1      & \cellcolor[rgb]{ .8,  1,  .6}0.8333 &
				\cellcolor[rgb]{ .792,  .929,  .984}1      & \cellcolor[rgb]{ .969,  .78,  .675}1      & \cellcolor[rgb]{ .8,  1,  .6}1	  &
				\cellcolor[rgb]{ .792,  .929,  .984}1      & \cellcolor[rgb]{ .969,  .78,  .675}1      & \cellcolor[rgb]{ .8,  1,  .6}1      &
				\cellcolor[rgb]{ .792,  .929,  .984}0.8000 & \cellcolor[rgb]{ .969,  .78,  .675}0.8000 & \cellcolor[rgb]{ .8,  1,  .6}0.8000 &  5 \\                                    

			3 & \cellcolor[rgb]{ .792,  .929,  .984}1      & \cellcolor[rgb]{ .969,  .78,  .675}0.9048 & \cellcolor[rgb]{ .8,  1,  .6}0.9500   & 
				\cellcolor[rgb]{ .792,  .929,  .984}1      & \cellcolor[rgb]{ .969,  .78,  .675}1      & \cellcolor[rgb]{ .8,  1,  .6}1      &
				\cellcolor[rgb]{ .792,  .929,  .984}0.9545 & \cellcolor[rgb]{ .969,  .78,  .675}1      & \cellcolor[rgb]{ .8,  1,  .6}0.9767 & 			
				\cellcolor[rgb]{ .792,  .929,  .984}0.9524 & \cellcolor[rgb]{ .969,  .78,  .675}0.9524 & \cellcolor[rgb]{ .8,  1,  .6}0.9524 & 21\\                                   	
      		4 & \cellcolor[rgb]{ .792,  .929,  .984}0.7111 & \cellcolor[rgb]{ .969,  .78,  .675}1      & \cellcolor[rgb]{ .8,  1,  .6}0.8312 & 
				\cellcolor[rgb]{ .792,  .929,  .984}0.7692 & \cellcolor[rgb]{ .969,  .78,  .675}0.9375 & \cellcolor[rgb]{ .8,  1,  .6}0.8451 &
				\cellcolor[rgb]{ .792,  .929,  .984}0.8750 & \cellcolor[rgb]{ .969,  .78,  .675}0.8750 & \cellcolor[rgb]{ .8,  1,  .6}0.8750 & 
				
				\cellcolor[rgb]{ .792,  .929,  .984}0.9000 & \cellcolor[rgb]{ .969,  .78,  .675}0.8438 & \cellcolor[rgb]{ .8,  1,  .6}0.8710 & 32\\                                   
			
			5 & \cellcolor[rgb]{ .792,  .929,  .984}1      & \cellcolor[rgb]{ .969,  .78,  .675}0.5288 & \cellcolor[rgb]{ .8,  1,  .6}0.6918 & 
				\cellcolor[rgb]{ .792,  .929,  .984}1      & \cellcolor[rgb]{ .969,  .78,  .675}0.5385 & \cellcolor[rgb]{ .8,  1,  .6}0.7000 &
				\cellcolor[rgb]{ .792,  .929,  .984}1      & \cellcolor[rgb]{ .969,  .78,  .675}0.5288 & \cellcolor[rgb]{ .8,  1,  .6}0.6918 & 
				 
				\cellcolor[rgb]{ .792,  .929,  .984}1      & \cellcolor[rgb]{ .969,  .78,  .675}0.5481 & \cellcolor[rgb]{ .8,  1,  .6}0.7081 & 104\\                                  
			
			A & 	\cellcolor[rgb]{ .792,  .929,  .984}       & \cellcolor[rgb]{ .969,  .78,  .675}       & \cellcolor[rgb]{ .8,  1,  .6}0.8715 & 
				\cellcolor[rgb]{ .792,  .929,  .984}       & \cellcolor[rgb]{ .969,  .78,  .675}       & \cellcolor[rgb]{ .8,  1,  .6}0.8808 & 
				\cellcolor[rgb]{ .792,  .929,  .984}       & \cellcolor[rgb]{ .969,  .78,  .675}       & \cellcolor[rgb]{ .8,  1,  .6}0.8762 & 
				\cellcolor[rgb]{ .792,  .929,  .984}       & \cellcolor[rgb]{ .969,  .78,  .675}       & \cellcolor[rgb]{ .8,  1,  .6}0.8715 & 428 \\                                  
			
			M & \cellcolor[rgb]{ .792,  .929,  .984}0.8725 & \cellcolor[rgb]{ .969,  .78,  .675}0.9018 & \cellcolor[rgb]{ .8,  1,  .6}0.8653 & 
				\cellcolor[rgb]{ .792,  .929,  .984}0.9290 & \cellcolor[rgb]{ .969,  .78,  .675}0.9117 & \cellcolor[rgb]{ .8,  1,  .6}0.9058 &
				\cellcolor[rgb]{ .792,  .929,  .984}0.9425 & \cellcolor[rgb]{ .969,  .78,  .675}0.9006 & \cellcolor[rgb]{ .8,  1,  .6}0.9072 &
				 
				\cellcolor[rgb]{ .792,  .929,  .984}0.9047 & \cellcolor[rgb]{ .969,  .78,  .675}0.8564 & \cellcolor[rgb]{ .8,  1,  .6}0.8672 & 428\\                                  
			
			W & \cellcolor[rgb]{ .792,  .929,  .984}0.8976 & \cellcolor[rgb]{ .969,  .78,  .675}0.8715 & \cellcolor[rgb]{ .8,  1,  .6}0.8615 & 
				\cellcolor[rgb]{ .792,  .929,  .984}0.9033 & \cellcolor[rgb]{ .969,  .78,  .675}0.8808 & \cellcolor[rgb]{ .8,  1,  .6}0.8706 & 
				\cellcolor[rgb]{ .792,  .929,  .984}0.9013 & \cellcolor[rgb]{ .969,  .78,  .675}0.8762 & \cellcolor[rgb]{ .8,  1,  .6}0.8656 & 
				
				\cellcolor[rgb]{ .792,  .929,  .984}0.8967 & \cellcolor[rgb]{ .969,  .78,  .675}0.8715 & \cellcolor[rgb]{ .8,  1,  .6}0.8625 & 428                                  \\
			\bottomrule
		\end{tabularx}%
		\begin{tablenotes}
			\scriptsize
			\item[$\bullet$]  C: Class; S:Support; A: Accuracy; M: Macro Average F1; W: Weighted Average F1;
			\item[$\bullet$]  For each model, columns are precision, recall and F1-score, respectively.
		\end{tablenotes}
	\end{threeparttable}
	\vspace{-10pt}
	\end{table}
\textbf{Overall Performance:}
All models exhibited relatively stable accuracy and F1 scores, with variations largely attributed to the dataset's class imbalance. The D-LSTM model consistently achieved the highest overall performance, highlighting the benefit of modeling temporal intervals via duration-based pseudo-embedding. This aligns with established findings that temporal regularities in healthcare data can improve model discriminability, especially when event durations carry semantic weight. In contrast, the DC-LSTM, while retaining some performance gains, demonstrated a drop in both accuracy and F1. This may be due to increased feature space complexity introduced by correlation pseudo-embedding, which likely amplified noise from dummy-coded variables and interfered with optimal hyperparameter tuning. The added redundancy may have diluted signal quality rather than enriching it. T-LSTM, which leverages label embeddings, achieved the third-best F1 score. Its moderate performance suggests that textual embeddings alone are insufficient in capturing event sequence dynamics in structured clinical data, particularly when not paired with temporal modeling. Moreover, the combined use of correlation and text embeddings in T-LSTM may have introduced incompatible representation biases. These observations indicate that simpler, targeted augmentations like duration embedding offer more robust generalization than multifaceted, high-dimensional combinations. Finally, B-LSTM showed the lowest performance, reaffirming the critical role of temporal and structural input augmentations in outcome prediction tasks involving heterogeneous event attributes.

\textbf{Class-Specific Performance:}
Per-class metrics reveal strong model performance on the majority classes (0-4), with recall exceeding 0.8 across all models and 55\% of these classes achieving perfect recall (1.0). Class 0, with a median range frequency (92 instances), achieved perfect precision, recall, and F1 across models—highlighting the models’ strong inductive bias toward well-represented patterns. Conversely, class 5 consistently showed low recall (0.52–0.55) and significant variability, indicating systemic challenges. This behavior can be attributed to three interrelated issues: (1) training noise or label ambiguity, which undermines the model’s confidence; (2) feature overlap with class 1 and class 4, which likely collapses decision boundaries during learning; and (3) distributional shift between training and testing splits, where efforts to generalize to minority classes may impair majority class performance due to the optimization of weighted F1 score. The consistent misclassification of 40–45 class 5 samples as class 1 corroborates this explanation and directly contributes to class 1’s inflated recall and reduced precision (0.78–0.81). Class 2, with only five samples, underscores the limitations of deep models under extreme data sparsity. T-LSTM and B-LSTM in particular fail to generalize to this class, likely due to insufficient representation learning at that level of class support. Interestingly, classes 3 and 4—also minority classes—still achieved reasonable performance, especially under D-LSTM. This suggests that temporal granularity via duration embeddings may help amplify weaker signals and stabilize learning for less frequent outcomes. Still, frequent confusion between classes 5 and 4 contributes to reduced precision in both, indicating a need for stronger boundary differentiation or class-specific regularization strategies.
\subsubsection{Simultaneous Outcome Prediction Results}
Table \ref{tab:result} summarizes the final performance of the best-tuned LSTM architectures on the BPIC12 and A/O sequences. The results are reported in terms of classification accuracy and benchmarked against prior research to highlight comparative effectiveness.
\begin{table}[htb]
\centering
\begin{threeparttable}
\setlength{\tabcolsep}{1pt}
\caption{Accuracy Scores of LSTM HyperModels and Previous Research Models on BPIC12 Dataset}
\label{tab:result}
\notesize
\begin{tabularx}{\linewidth}{llllllllllX}
\hline
 & SVM\cite{teinemaa2019outcome} & LR\cite{teinemaa2019outcome} & RF\cite{teinemaa2019outcome} & XGB\cite{teinemaa2019outcome} & LSTM\cite{wang2019outcome} & CNN\cite{pasquadibisceglie2020orange} & DT\cite{donadello2023outcome}   & \textbf{M$^\dagger$} & \textbf{F$^\dagger$} & \textbf{U$^\dagger$} \\
    accept & 0.63 & 0.65 & 0.69 & 0.7  & 0.71 & 0.67 & 1    & \textbf{1}    & \textbf{1}    & \textbf{1} \\
    decline & 0.55 & 0.59 & 0.6  & 0.62 & 0.64 & 0.61 & 1    & \textbf{1}    & \textbf{1}    & \textbf{1} \\
    cancel & 0.70 & 0.69 & 0.7  & 0.7  & 0.73 & 0.7  & 1    & \textbf{1}    & \textbf{1}    & \textbf{1} \\
    avg & 0.63 & 0.64 & 0.66 & 0.67 & 0.69 & 0.66 & 1    & \textbf{1}    & \textbf{1}    & \textbf{1} \\
\bottomrule
\end{tabularx}
    \begin{tablenotes}[flushleft]
	\scriptsize
	\item M: M-B-LSTM; F: F-B-LSTM; U:F-D-LSTM
	\item M-B-LSTM and F-B-LSTM accuracy score for BPI12A/O are $1$.
    \end{tablenotes}
\end{threeparttable}
\end{table}

Previous studies~\cite{wang2019outcome,weytjens2020process} approached the simultaneous outcome prediction task by decomposing it into three separate binary classification problems, requiring the training of three distinct models per instance. In contrast, our LSTM HyperModels employ a single multiclass classifier to distinguish among the \textit{accept}, \textit{decline}, and \textit{cancel} outcomes. Despite the simpler architecture, our models achieve perfect accuracy (100\%) across all three classes on the BPIC12 and A/O variants, demonstrating both high predictive performance and computational efficiency. This level of performance aligns with or surpasses prior results—such as the decision tree ensemble~\cite{donadello2023outcome}—while avoiding the complexity of multi-model setups.

We attribute this success to our proposed multidimensional embedding strategy and the incorporation of time-difference flagging and duration pseudo-embedding matrix, which together enhance the model’s ability to capture subtle event dynamics. By consolidating the task into a single classifier, our approach significantly reduces training and inference overhead, making it well-suited for real-time applications in business process monitoring.

However, the simplicity of the BPIC12-A/O datasets may amplify this effectiveness. Many events co-occur in fixed, repetitive patterns, resulting in reduced sequence diversity. This structural regularity likely simplifies the classification task and may inflate performance on such benchmarks. To assess robustness, future evaluations should test generalizability on datasets with greater variability, noise, or temporal irregularity.

Overall, these findings underscore the strength of our encoding strategies in capturing fine-grained temporal and relational dynamics in simultaneous outcome prediction. At the same time, they highlight the need for broader validation to ensure the approach remains effective in more complex, high-entropy process environments.

\section{Conclusion and Discussion}
\label{sec:CC}
This paper presents a framework for outcome prediction in predictive business process monitoring (PBPM), integrating LSTM-based HyperModels with advanced attribute encoding and embedding strategies. We introduce novel methods such as pseudo-embedding for universal attributes, duration-based binning, and multidimensional embeddings with time-difference flag augmentation, designed to handle challenges in simultaneous event prediction. Addressing the gap in unified outcome prediction frameworks, our contribution lies in a flexible toolkit for different task modeling, enriching event representation through novel embedding strategies, and validating performance across multiple domains. This provides a practical and theoretically sound pathway for scalable PBPM in critical application areas such as healthcare, logistics, and enterprise operations.

Experiments across four datasets—Patients, BPIC12, and BPIC12-A/O—demonstrate the efficacy of these techniques in both imbalanced and balanced scenarios. The models, leveraging event label processing and pseudo-embedding, effectively capture complex relationships and temporal dependencies. Multiple LSTM architectures offer flexibility for different prediction tasks, with dynamic hyperparameter optimization ensuring robustness across diverse datasets.

Results show significant improvements in predictive accuracy, showcasing the power of the proposed strategies in capturing event attribute interplay and temporal dynamics. This work not only advances outcome-oriented sequence modeling but also provides a scalable solution for real-world PBPM applications.

Despite promising results, the proposed framework faces several limitations. First, the self-tuning process introduces non-trivial computational overhead, particularly in large-scale or time-sensitive applications. Second, while the current evaluation shows strong results on structured datasets, further validation is needed to ensure generalizability to domains with irregular sampling, missing data, or high noise. Third, although the architecture is aligned with the encoding logic, domain-specific hybrid optimization strategies—such as incorporating expert knowledge or meta-learning approaches—could further improve convergence and interpretability. Fourth, while the time-difference flag supports temporal ordering, future work should include controlled ablation studies to assess its standalone contribution to performance.

Going forward, our roadmap includes expanding the framework to accommodate concurrent event streams, next-event prediction, and full sequence modeling. The M-B-LSTM and F-B-LSTM variants are well-positioned for these tasks, given their multidimensional embeddings and context-aware structure. Moreover, integrating these models with transformer architectures or attention mechanisms may enhance the ability to capture long-range dependencies. Another future extension is to integrate uncertainty estimation, especially in noisy or low-data scenarios.

Finally, deployment in real-world business process monitoring platforms remains a critical milestone. Case studies in verticals such as healthcare, logistics, and finance could offer valuable insights into human-in-the-loop interaction, interpretability requirements, and system integration challenges. These efforts will be essential to transition the framework from experimental validation to impactful, domain-specific applications.

\bibliographystyle{IEEEtran}
\bibliography{manuscript}

\vfill

\end{document}